\documentclass[10pt,twocolumn,letterpaper]{article}

\usepackage{wacv}              % To produce the CAMERA-READY version
\usepackage{graphicx}
\usepackage{amsmath}
\usepackage{amssymb}
\usepackage{booktabs}
\usepackage{reledmac}
\usepackage{subcaption}
\usepackage{float}
\usepackage[english]{babel}
\usepackage{csquotes}

\usepackage[pagebackref,breaklinks,colorlinks]{hyperref}

\usepackage[capitalize]{cleveref}
\crefname{section}{Sec.}{Secs.}
\Crefname{section}{Section}{Sections}
\Crefname{table}{Table}{Tables}
\crefname{table}{Tab.}{Tabs.}

\def\wacvPaperID{*****} % *** Enter the WACV Paper ID here
\def\confName{WACV}
\def\confYear{2025}

\begin{document}

\title{SGNetPose+: Stepwise Goal-Driven Networks with Pose Information for Trajectory Prediction in Autonomous Driving}

\author{
    {\large \textbf{Akshat Ghiya, Ali K. AlShami, Jugal Kalita}}\\[0.5em] % Larger, bold names, with extra space
    University of Colorado Colorado Springs\\
    1420 Austin Bluffs Pkwy, Colorado Springs, CO 80918, USA\\
    {\tt\small aghiya@uccs.edu, aalshami@uccs.edu, jkalita@uccs.edu}
}

\maketitle

%%%%%%%%% ABSTRACT
\begin{abstract}

Predicting pedestrian trajectories is essential for autonomous driving systems, as it significantly enhances safety and supports informed decision-making. Accurate predictions enable the prevention of collisions, anticipation of crossing intent, and improved overall system efficiency. In this study, we present SGNetPose+, an enhancement of the SGNet architecture designed to integrate skeleton information or body segment angles with bounding boxes to predict pedestrian trajectories from video data to avoid hazards in autonomous driving. Skeleton information was extracted using a pose estimation model, and joint angles were computed based on the extracted joint data. We also apply temporal data augmentation by horizontally flipping video frames to increase the dataset size and improve performance. Our approach achieves state-of-the-art results on the JAAD and PIE datasets using pose data with the bounding boxes, outperforming the SGNet model. Code is available on \href{https://github.com/Aghiya/SGNetPose}{Github: SGNetPose+}.

\end{abstract}

%%%%%%%%% BODY TEXT
\section{Introduction}
\label{sec:intro}
The number of increasingly autonomous cars on the market has increased significantly in recent years. People are increasingly choosing such vehicles for greater comfort and safety, aiming to reduce human error, a leading cause of road accidents worldwide. Advanced technologies in Autonomous Driving Systems, such as sensors, cameras, and LiDAR, allow these systems to detect and respond to hazards with remarkable precision and speed. These new systems have the potential to save lives by reducing collisions and creating safer roads for drivers, passengers, and pedestrians alike.

\begin{figure}
    \centering
    \includegraphics[width=0.9\linewidth]{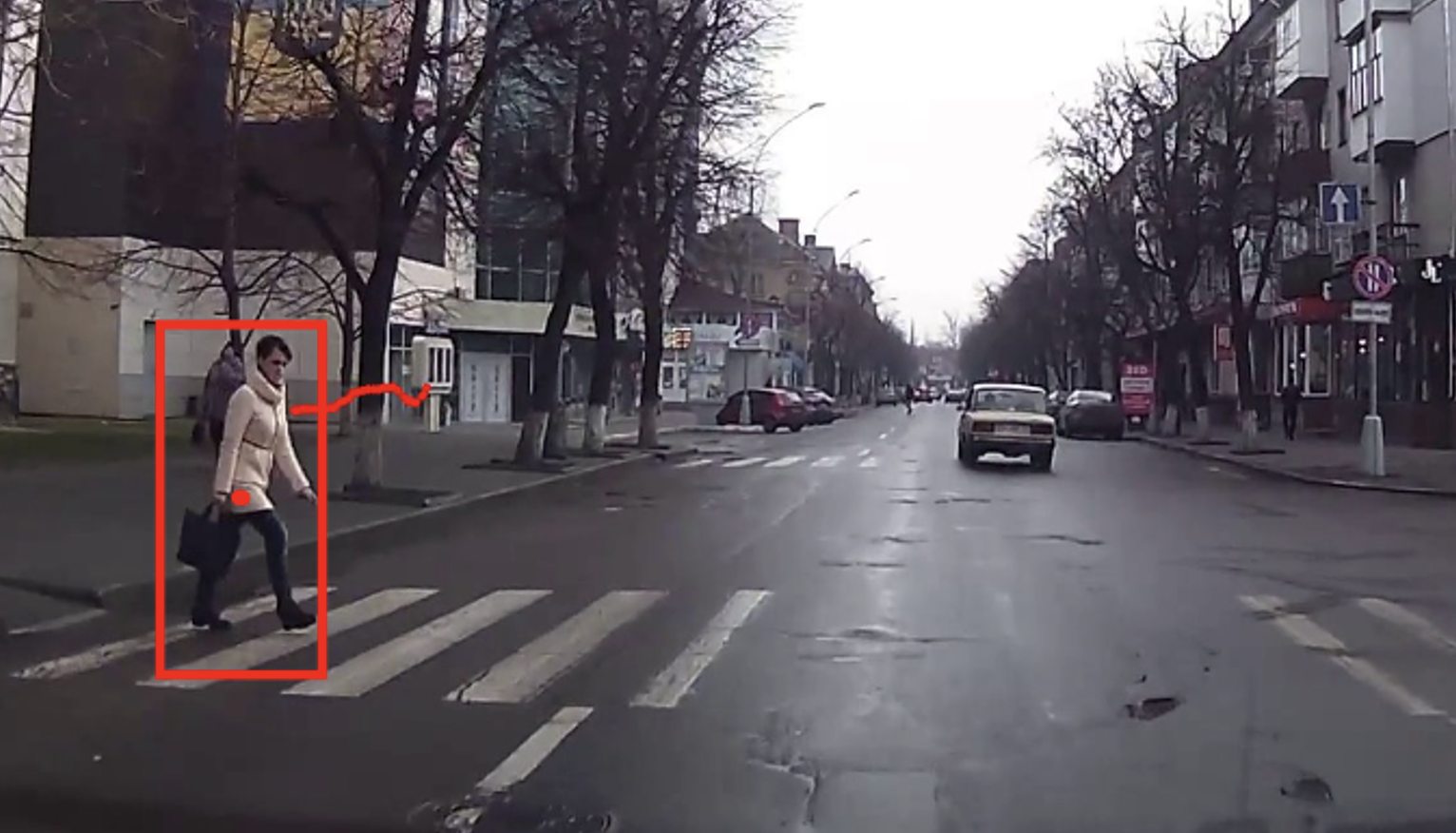}
    \caption{A frame from the JAAD dataset, showing a woman in a white coat crossing the street with her bounding box superimposed. A car in approaching in the center lane to her right.}
    \label{fig:enter-label}
\end{figure}

Safety remains the cornerstone of autonomous driving technology~\cite{alshami2024coool}. Over half of global road fatalities involve vulnerable road users, such as pedestrians, cyclists, and motorcyclists \cite{zhang2020research}. The Global Status Report on Road Safety 2023 highlights that nearly 1.19 million people lost their lives in automobile accidents \cite{WHO2023RoadSafety}. Advancing the safety of autonomous systems is crucial to saving lives and protecting drivers, passengers, and pedestrians alike.

Predicting pedestrian trajectories is essential for ensuring safety in autonomous driving systems. Knowing where pedestrians are going and predicting their intention to cross is crucial to avoiding accidents. However, this task is challenging. Pedestrians are highly agile and can unpredictably change their speed and direction, often displaying complex and hard-to-predict movement patterns \cite{zhang2020research}. Additionally, their behavior is influenced by the surrounding context. For example, a pedestrian walking on the sidewalk might suddenly cross the street outside a designated crosswalk if they fail to notice an oncoming car. Despite such rule violations, vehicles and their drivers are more responsible for avoiding harm to pedestrians than the pedestrians themselves.

While bounding boxes detect and track pedestrians, they provide a coarse representation that overlooks critical details such as movement dynamics that can be inferred from the body posture. Skeleton data, which capture the spatial-temporal configuration and joint orientations of the human body, enables the model to recognize subtle motion cues~\cite{alshami2024smart}, like a shift in weight or an extended leg, that indicate specific actions or intentions, including turning, accelerating, or stopping. This detailed understanding proves invaluable in complex environments where accurate trajectory prediction must consider diverse interactions, obstacles, and social behaviors.

In this work, we propose SGNetPose+, an enhanced version of the SGNet model that predicts pedestrian trajectories by leveraging bounding boxes with skeleton joint information or the angle of the body joints using two encoders rather than relying only on bounding boxes. The ViTPose human pose estimation model is utilized to extract skeleton data. Some of our experiments use body angle information with and without skeleton data calculated from skeleton joints. Additionally, we apply temporal data augmentation for the JAAD dataset by horizontally flipping video frames to increase the data and enhance the results, offsetting the dataset size after filtering for frames where ViTPose successfully inferred skeleton points. Our encoder and decoder architecture is RNN-based. While transformers have been used to predict trajectories \cite{Li_2022_CVPR, Shi_2023_ICCV}, it is reported that they do better with larger datasets \cite{LIN2022111}. Our datasets are not necessarily large enough to clearly justify their use. Our main contributions are as follows:

\begin{itemize}
\item We introduce a novel architecture, SGNetPose+, designed to effectively capture complex spatial patterns and joint relationships by leveraging bounding box data, pose information using two encoders.
\item We present $JAAD_{pose}$ and $PIE_{pose}$, enhanced subsets of the original datasets enriched with pose information, body segment angles, and temporal augmentation for predicting future trajectories. Pose extraction using ViTPose and temporal augmentation applied by flipping the image with the coordinates horizontally. The datasets are available on \href{https://github.com/Aghiya/SGNetPose}{Github:~$JAAD_{pose}$, $PIE_{pose}$}.
\item Our method achieved better performance compared to the baseline model, SGNet, in predicting pedestrian trajectories on the $JAAD_{pose}$ and $PIE_{pose}$ datasets using pose data with temporal augmentation.
\end{itemize}

In the following sections, we discuss papers with connected work, our approach and the architecture for SGNetPose+. Lastly, we discuss our experimental results and their meaning.

\section{Related Work}
Pedestrian trajectory prediction focuses on forecasting the path of pedestrians by representing sequences of centroids or bounding box coordinates. Due to the sequential nature of this task, many approaches leverage recurrent neural networks (RNNs), such as LSTMs \cite{bock2017self, saleh2017intent, saleh2018intent} and GRUs \cite{liu2020spatiotemporal, rasouli2017they}. Convolutional Neural Networks (CNNs) are also commonly used, particularly when input data consists of video frames captured from bird's-eye or vehicle-mounted perspectives. For instance, Liang et al. \cite{liang2019peeking} utilize CNNs to extract spatial features, which are then passed to Conv-LSTMs to capture temporal features for improved trajectory prediction. Some studies showed promising results utilizing Transformer models for pedestrian trajectory prediction~\cite{yin2021multimodal}, while others have applied them to predict player trajectories in sports~\cite{alshami2023pose2trajectory}.

Predicting pedestrian trajectories can complement other approaches, such as crossing intention prediction, which determines whether a pedestrian will cross the street—an essential task for vehicle decision-making. Models for crossing intention prediction often utilize sequential data and RNN-based architectures, as demonstrated in \cite{piccoli2020fussi, zhang2020research}.

To enhance prediction performance, researchers have incorporated additional data beyond conventional motion information, including pedestrian pose, road rules, camera motion, and social forces. For example, Lorenzo et al. \cite{lorenzo2020rnn} utilize CNNs to extract pedestrian appearance features for crossing intention prediction, while Kotseruba et al. \cite{kotseruba2021benchmark} include skeletal posture information. Additionally, social interactions between pedestrians have been explored; Fernando et al. \cite{fernando2018soft} propose social pooling layers to model these dynamics, significantly improving trajectory prediction accuracy.

\begin{figure*}{}
    \centering
    \includegraphics[width=0.8\linewidth]{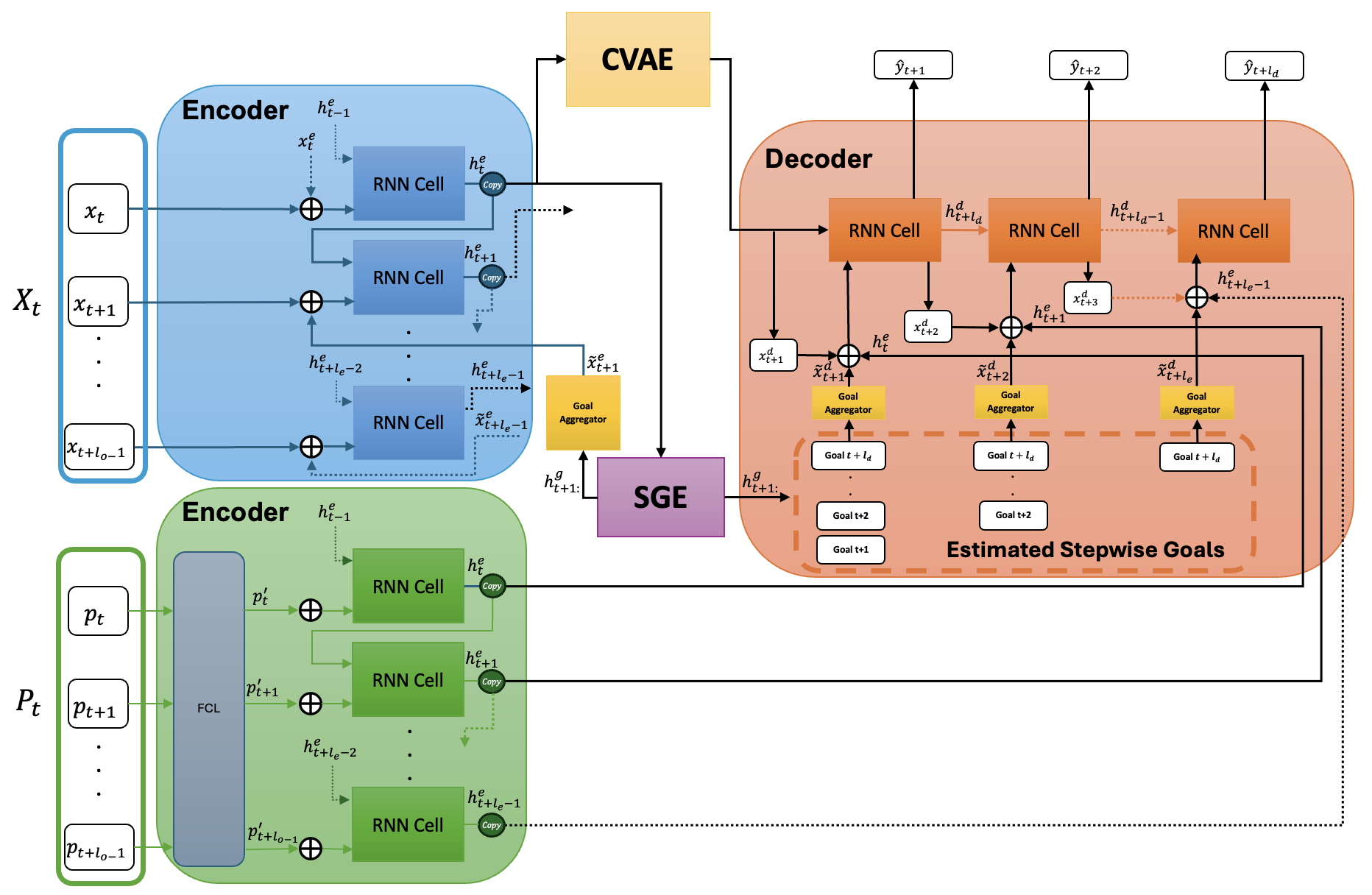}
    \caption{Visualization of SGNetPose+. Encoder time evolves vertically, from time \( t \) to \( t+1 \), while decoder time flows horizontally, predicting the trajectory from \( t+1 \) to \( t+ld \). At the start, trajectory information (bounding boxes, denoted as $x^t$) are fed to an RNN cell, while the pose information (skeleton or body angle, denoted as $p^t$) are fed to a separate RNN cell. The output of the cell is fed to the CVAE and to the SGE (Stepwise Goal Estimator). The CVAE output, SGE's estimated goals, and RNN cell outputs are combined and fed to the decoder, which produces the predicted location $y^t$.}
    \label{fig:SGNetPose}
\end{figure*}

Traditionally, trajectory prediction and crossing intention were treated as separate tasks, each requiring its own model. However, recent research has shown the potential and advantages of addressing these tasks jointly. Studies such as those by Huang et al. \cite{huang2020long} and Goldhammer et al. \cite{goldhammer2019intentions} demonstrate that combining trajectory and intention modeling enhances predictive performance, underscoring the interconnected nature of these problems. Our work focuses on using skeleton and body angle information for trajectory prediction, which is not traditionally done.

\section{Approach}

As shown in Figure~\ref{fig:SGNetPose}, our SGNetPose+ architecture is designed to predict pedestrian trajectories by leveraging both bounding box and pose information through two encoders, utilizing 13 key points for skeleton representation and two for bounding box data. Each skeleton joint and bounding box point is represented by its $x$ and $y$ coordinates, corresponding to the pixel values within the frame. The first encoder processes bounding box data, capturing historical information alongside predicted stepwise goals to enrich the hidden representation, enhancing future predictions and generating new stepwise goals for subsequent time steps. The second encoder handles pose inputs, passing them through a fully connected (FC) layer with a dropout layer to obtain embeddings, which are then fed to the RNN. The RNN outputs are subsequently sent to the decoder for further processing. The decoder leverages the stepwise goals from the SGE, CVAE, and pose encoder outputs to predict future trajectories effectively. A Conditional Variational Autoencoder (CVAE) is employed for latent goal modeling, while a Stepwise Goal Estimation (SGE) module is utilized for iterative trajectory prediction. CVAE and SGE are built upon the SGNet baseline following other prior works such as Bitrap~\cite{yao2021bitrap} and PECNet~\cite{mangalam2020not}. The following sections detail the mathematical framework and the innovations introduced in SGNetPose+.

\subsection{Input and Output Representation}
Let $X_t = \{x_t, x_{t+1}, \dots, x_{t+L_o-1}\}$ denote the observed bounding box sequence over $L_o$ frames, where each $x_t \in \mathbb{R}^4$ represents a bounding box as $[x_{\text{min}}, y_{\text{min}}, x_{\text{max}}, y_{\text{max}}]$. Simultaneously, $P_t = \{p_t, p_{t+1}, \dots, p_{t+L_o-1}\}$ represents the observed pose sequence, where each pose $p_t \in \mathbb{R}^{J \times 2}$ corresponds to $J$ keypoints with 2D coordinates. The goal is to predict the future trajectory of the pedestrian, represented as $Y_t = \{\hat{y}_{t+1}, \hat{y}_{t+2}, \dots, \hat{y}_{t+l_d}\}$, where $\hat{y}_{t+k} \in \mathbb{R}^2$ in the next $l_d$ frames.

\subsection{Encoding Observations}
The first encoder captures an agent’s movement behavior from the bounding boxes as a latent vector by embedding its historical trajectory $X_t$. The input feature $x^e_{t}$ is then concatenated with the aggregated goal information $\tilde{x}^e_{t}$ from the previous time step $t{-}1$, and then the new hidden state $h^e_{t}$ is updated through a recurrent cell. Hidden state $h^e_{t+1}$ and $x^e_{t+1}$ are both set to zero for the first time step. The second encoder captures an agent's movement behavior from pose data as a latent vector by embedding its historical trajectory $P_t$ through a single fully connected layer. The output of the bounding box encoder is fed to the CVAE to model latent goals, while the output of the pose encoder is used later in the decoder by concatenating it with other components.

\subsection{Conditional Variational Autoencoder (CVAE)}
To model the inherent uncertainty in pedestrian motion, SGNetPose+ works the same as SGNet, which employs a Conditional Variational Autoencoder (CVAE) framework, which is applied in the SGNet model following prior works such as Bitrap~\cite{yao2021bitrap} and PECNet~\cite{mangalam2020not}. The CVAE framework learns the distribution of the future trajectory $Y_t$ conditioned on the observed trajectory $X_t$ by introducing a latent variable $z$.

During training, the ground truth future trajectory $Y_t$ is fed into the target encoder, which outputs the hidden state $h_{Y_t}$. To capture the dependencies between the observed and ground truth trajectories, the recognition network utilizes the hidden states $h^e_t$ and $h_{Y_t}$ to predict the distribution mean $\mu^q_z$ and standard deviation $\sigma^q_z$. A sample from $z \sim \mathcal{N}(\mu^q_z, \sigma^q_z)$ is then concatenated with $h^e_t$ to generate $h^d_t$ through the generation network.

During testing, the ground truth future trajectory is not available. Instead, the generation network concatenates $z$, sampled from $\mathcal{N}(\mu^p_z, \sigma^p_z)$, with $h^e_t$ to produce $h^d_t$.

\subsection{Stepwise Goal Estimator (SGE)}

The concept of SGE involves generating coarse, stepwise goals to guide trajectory prediction through a hierarchical, coarse-to-fine framework. These stepwise goals also support the generation of subsequent goals for future steps, ensuring continuity in the prediction process. To implement this, SGE predicts and propagates the coarse stepwise goals to both the encoder and decoder.

In the encoder, at each time step $t$, the stepwise goals from the previous time step ($t{-}1$) are concatenated with the input features to provide supplementary context. This additional information enables the encoder to learn a more discriminative representation of the sequence. To address the potential issue of inaccurate future goals misguiding the predictions, a goal aggregation module is employed. This module uses an attention mechanism to adaptively evaluate and assign importance to each stepwise goal, ensuring that the model prioritizes the most relevant goals and improves prediction reliability. Goal Aggregator for Encoder and Decoder are defined as following:

    \begin{equation}
        w = \text{Softmax}(W^T \text{Tanh}(h_{t+i}^g) + b),
    \end{equation}
    and the aggregated goal $\tilde{x}_{t+i}$ is given by:
    \begin{equation}
        \tilde{x}_{t+i} = f_{\text{attn}}(h_{t+i}^g) = \sum_{s=t+i}^{\ell_d} w_s h_s^g.
    \end{equation}

\subsection{Decoder Integration}  
As previously mentioned, the recurrent decoder generates the final trajectory \(Y_t = \{\hat{y}_{t+1}, \hat{y}_{t+2}, \dots, \hat{y}_{t+l_d}\}\) using a trajectory regressor. At each time step, given the hidden state \(h^d_t\)\ from the bounding box and pose encoders and the estimated goal input \(\tilde{x}^d_{t+i}\), the decoder updates its hidden state for the next step through a recurrent cell. The trajectory regressor, implemented as a single fully-connected layer, takes the hidden states \(h^d_{t+i}\) and computes the predicted trajectory \(\hat{Y}_{t+i}\) for each time step. This design enables SGNetPose+ to outperform baseline models in trajectory prediction tasks, especially under complex and dynamic environments.

\subsection{Datasets}

We used two datasets: the Joint Attention in Autonomous Driving (JAAD) dataset \cite{rasouli2017they} and the Pedestrian Intention Estimation (PIE) dataset \cite{rasouli2019pie}. JAAD consists of 246 videos taken from an on-board camera and are between five to ten seconds long. Each frame is annotated with pedestrian bounding boxes, current pedestrian and vehicle action, and more. PIE consists of 300,000 labeled video frames taken from an on-board camera and includes vehicle speed, location, etc.

To extract pedestrian skeletons, we employed a pre-trained version of the Vision Transformer model, ViTPose \cite{xu2022vitpose}, which generated 2D skeletons comprising 13 key points. These points included the nose, shoulders, elbows, wrists, hips, knees, and ankles. From these 13 key points, 12 angles were computed based on the methodology proposed by Sidharta et al. \cite{sidharta2024small}.

We evaluate our model on a subset of the JAAD and PIE datasets ($JAAD_{pose}$ and $PIE_{pose}$), where pose information has been used. The JAAD and PIE datasets provide trajectory bounding box annotations for all individuals in each video frame, including pedestrians attempting to cross the street and those on the sidewalk. However, due to the low quality of some bounding boxes, the pose estimation model struggled to accurately predict poses in many frames, resulting in a reduced dataset size. Despite this limitation, the primary goal of this study is to highlight the critical role pose information can play in enhancing trajectory prediction accuracy. Since the amount of JAAD data was reduced after extracting pose information, we applied temporal augmentation techniques to expand the dataset and incorporate more pose-inclusive examples. Detailed counts of the resulting dataset subsets are presented in Table \ref{table_data_counts}.

\begin{table}[ht]
\resizebox{\columnwidth}{!}{
\begin{tabular}{|c|c|c|c|c|}
\hline
\textbf{Dataset} & Augmentation &\textbf{Training} & \textbf{Validation} & \textbf{Testing}\\ \hline
JAAD (Original) & No & 200 & 34 & 161 \\ \hline
JAAD (Inferred) & No & 59 & 9 & 51 \\ \hline
JAAD (Augmented)& Yes & 118 & 18 &102 \\ \hline
PIE (Original) & No & 524 & 121 & 426 \\ \hline
PIE (Inferred) & No & 433 & 81 & 314 \\ \hline

\end{tabular}
 }
\caption{Number of batches generated from data (train/val/test). “Original" refers to the data used with baseline the SGNet. “Inferred" is the data extracted when restricted to requiring skeletal information for a pedestrian. “Augmented" is the data that includes the data we generated via augmentation.}
\label{table_data_counts}
\end{table}

The body segment angles were defined as follows: the neck angle was calculated using the nose, the midpoint of the shoulders, and the left or right shoulder. The armpit angle was determined using the elbow, the corresponding shoulder, and the shoulder-to-shoulder line. The elbow angle involved the wrist, the corresponding elbow, and the shoulder on the same side. The torso angle was computed using the hip, the corresponding shoulder, and the hip-to-knee line. The thigh angle utilized the hip, the corresponding knee, and the line between both hips. Finally, the knee angle was determined using the knee, the corresponding heel, and the hip. Once the key points and angles were computed, we extracted sequences from both datasets and organized them into a custom dataset format, enabling efficient loading for training purposes.

\begin{table*}[ht]
\centering
% To make the table span a percentage of the page width, use \resizebox
\resizebox{\textwidth}{!}{
\begin{tabular}{|c|c|c|c|c|c|c|c|}
\hline
\textbf{Model Details} & \textbf{Features} & \textbf{MSE 15 F/0.5s} & \textbf{MSE 30 F/1s} & \textbf{MSE 45 F/1.5s} & \textbf{FMSE} & \textbf{CMSE} & \textbf{CFMSE}\\ \hline
Base SGNet~\cite{wang2022stepwise} & BB & 68.46 & 157.31 & 384.25 & 1250.61 & 297.08 & 1050.27 \\ \hline
SGNetPose+ & BB +Pose & \textbf{62.62} & \textbf{146.65} & \textbf{347.07} & \textbf{1080.45} & \textbf{260.17} & \textbf{872.99} \\ \hline
SGNetPose+ & BB + Angle & 72.06 & 165.06 & 405.06 & 1319.83 & 308.60 & 1098.36 \\ \hline
\end{tabular}
}
\caption{Results on the $JAAD_{pose}$ Dataset \cite{rasouli2017they} for predicting pedestrian trajectories. The evaluation includes Mean Squared Error (MSE) for 15 frames (0.5 seconds), 30 frames (1 second), and 45 frames (1.5 seconds) ahead, along with Final MSE (FMSE), Cumulative MSE (CMSE), and Cumulative Final MSE (CFMSE). \textbf{Pose} refers to skeleton keypoints, \textbf{Angle} refers to the body segment angle, and \textbf{BB} refers to bounding boxes. SGNetPose+ with pose data shows notable improvements.}
\label{table_res_jaad}
\end{table*}

\begin{table*}[ht]
\centering
% To make the table span a percentage of the page width, use \resizebox
\resizebox{\textwidth}{!}{
\begin{tabular}{|c|c|c|c|c|c|c|c|}
\hline
\textbf{Model Details} & \textbf{Features}&\textbf{MSE 15 F/0.5s} & \textbf{MSE 30 F/1s} & \textbf{MSE 45 F/1.5s} & \textbf{FMSE} & \textbf{CMSE} & \textbf{CFMSE}\\ \hline
Base SGNet~\cite{wang2022stepwise} & BB &17.05 & 41.98 & 104.29 & 341.57 & 81.28 & 282.21 \\ \hline
SGNetPose+ & BB + Pose &\textbf{15.81} & \textbf{40.08} & \textbf{102.93} & 343.42 & 80.84 & 286.03 \\ \hline
SGNetPose+ & BB + Angle & 16.84 & 41.81 & 103.19 & \textbf{332.55} & \textbf{79.99} & \textbf{271.96} \\ \hline
\end{tabular}
}
\caption{Results on the $PIE_{pose}$ Dataset~\cite{rasouli2019pie}  for predicting pedestrian trajectories. Metrics include MSE for 15 frames (0.5 seconds), 30 frames (1 second), and 45 frames (1.5 seconds) ahead, along with Final MSE (FMSE), Cumulative MSE (CMSE), and Cumulative Final MSE (CFMSE). \textbf{Pose} refers to skeleton keypoints, \textbf{Angle} refers to the body segment angle, and \textbf{BB} refers to bounding boxes. SGNetPose+ with pose data shows notable improvements.}

\label{table_res_pie}
\end{table*}

\begin{figure}[H]
\centering
\begin{subfigure}[t]{0.45\textwidth}
    \centering
    \includegraphics[width=\textwidth]{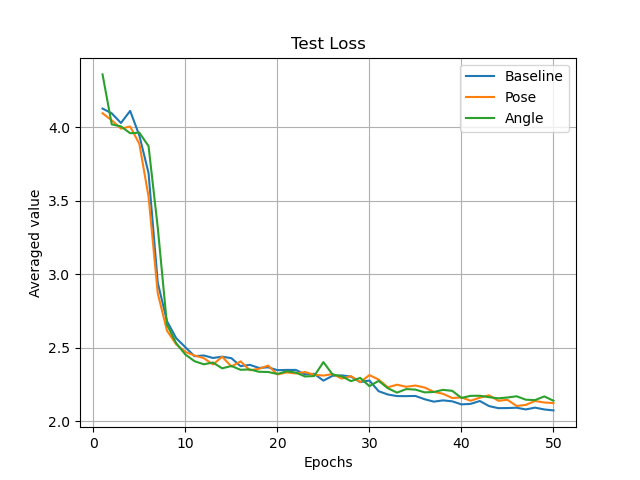} % Replace with your image file path
    \caption{Test Loss for $JAAD_{beh}$ (our reduced JAAD dataset with pose information) data using Root Mean Square Error (RMSE) Loss function}
    \label{fig:jaad_test_loss}
\end{subfigure}%
\hspace{0.05\textwidth} % Adjust horizontal spacing if needed
\begin{subfigure}[t]{0.45\textwidth}
    \centering
    \includegraphics[width=\textwidth]{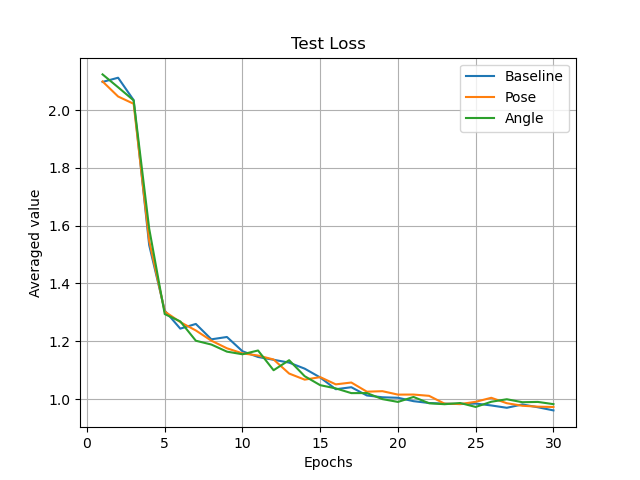} % Replace with your image file path
    \caption{Test Loss for $PIE_{beh}$ (our reduced PIE dataset with pose information) data using Root Mean Square Error (RMSE) Loss function}
    \label{fig:pie_test_loss}
\end{subfigure}
\caption{Comparison of test loss for JAAD and PIE datasets.}
\label{fig:comparison}
\end{figure}

\section{Experiments}

In our experiments, we trained and evaluated our novel architecture, SGNetPose+, which incorporates bounding box data alongside pose and body segment angle information, utilizing a dual-encoder and single-decoder design. The $JAAD_{pose}$ and $PIE_{pose}$ datasets, subsets of the original data enriched with pose and angle details, were utilized for training. To ensure a fair evaluation, the same dataset subsets used for training the SGNet~\cite{wang2022stepwise} baseline model were employed for comparison.

Various metrics were used to assess the performance of the model. Mean Square Error (MSE) was calculated to measure the average squared error between the predicted and ground truth bounding box coordinates. This was computed for 15, 30, and 45 frames into the future, corresponding to 0.5 seconds, 1.0 seconds, and 1.5 seconds, respectively, given the 30 FPS video frame rate. Final Mean Square Error (FMSE) was used to evaluate the average squared error specifically for the final predicted bounding boxes. Centroid Mean Square Error (CMSE) measured the average squared error for the top-left corner of the predicted bounding boxes, while Centroid Final Mean Square Error (CFMSE) focused on the top-left corner of the final predicted bounding boxes. These metrics provide a comprehensive evaluation of the model's accuracy in predicting future bounding box trajectories.

\begin{figure*}[ht]
    \centering
    % JAAD Figures
    \begin{subfigure}[t]{0.95\textwidth}
        \centering
        \includegraphics[width=\linewidth]{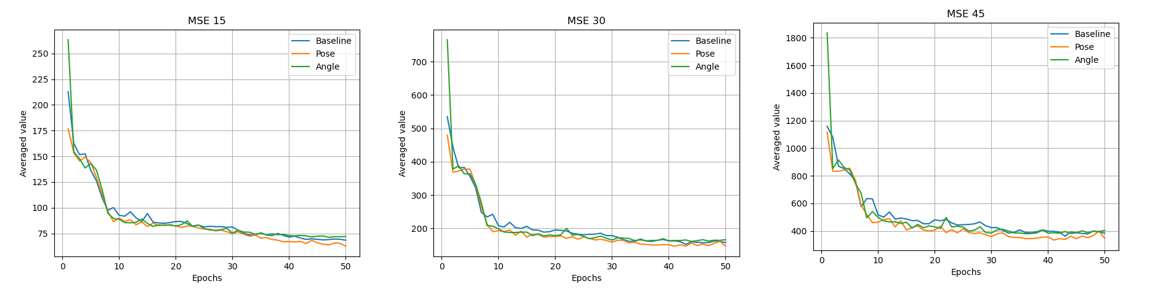}
        \caption{Evaluating the SGNetPose+ model on $JAAD_{pose}$ to predict pedestrian trajectory using the Mean Square Error (MSE) metric for predictions 15 frames (0.5 seconds), 30 frames (1 second), and 45 frames (1.5 seconds) into the future.}
        \label{fig:jaad_mse}
    \end{subfigure}
    \vskip\baselineskip
    \begin{subfigure}[t]{0.95\textwidth}
        \centering
        \includegraphics[width=\linewidth]{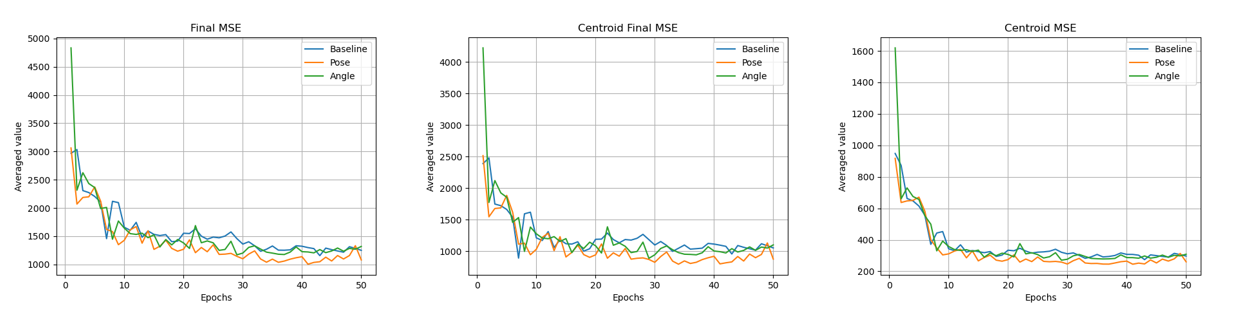}
        \caption{Evaluating the SGNetPose+ model on $JAAD_{pose}$ to predict pedestrian trajectory using different metrics, including Final MSE (FMSE), Cumulative MSE (CMSE), and Cumulative Final MSE (CFMSE).}
        \label{fig:jaad_other_metrics}
    \end{subfigure}
    \vskip\baselineskip
    % PIE Figures
    \begin{subfigure}[t]{0.95\textwidth}
        \centering
        \includegraphics[width=\linewidth]{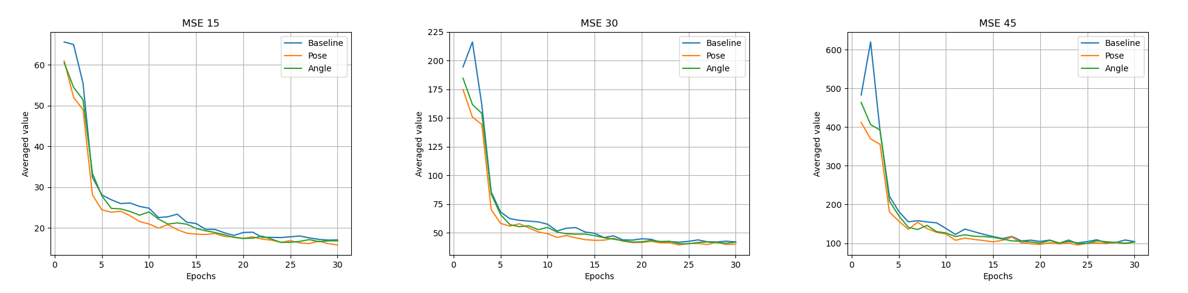}
        \caption{Evaluating the SGNetPose+ model on $PIE_{pose}$ to predict pedestrian trajectory using the Mean Square Error (MSE) metric for predictions 15 frames (0.5 seconds), 30 frames (1 second), and 45 frames (1.5 seconds) into the future.}
        \label{fig:pie_mse}
    \end{subfigure}
    \vskip\baselineskip
    \begin{subfigure}[t]{0.95\textwidth}
        \centering
        \includegraphics[width=\linewidth]{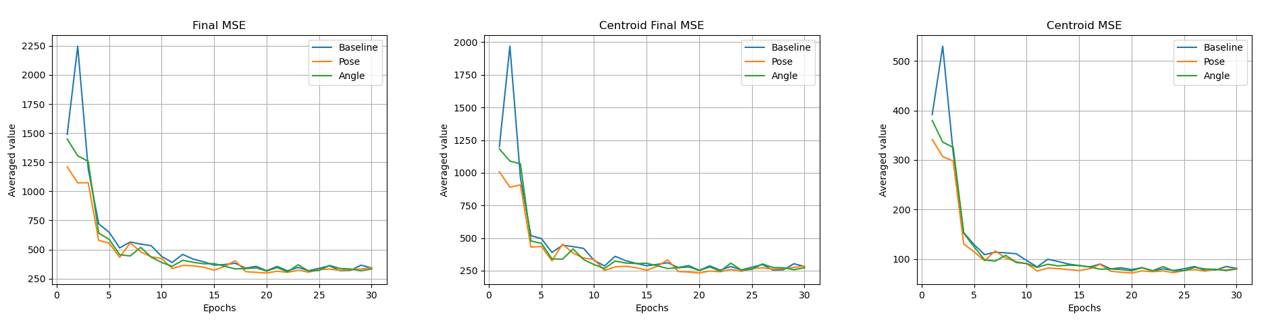}
        \caption{Evaluating the SGNetPose+ model on $PIE_{pose}$ to predict pedestrian trajectory using different metrics, including Final MSE (FMSE), Cumulative MSE (CMSE), and Cumulative Final MSE (CFMSE).}
        \label{fig:pie_other_metrics}
    \end{subfigure}
    \caption{Comparison of $JAAD_{pose}$ and $PIE_{pose}$ with different metrics.}
    \label{fig:combined_jaad_pie}
\end{figure*}

\section{Results}
Tables \ref{table_res_jaad} and \ref{table_res_pie} present the results of our model compared to the baseline on the $JAAD_{pose}$ and $PIE_{pose}$ datasets. Figures \ref{fig:jaad_test_loss} illustrate the averaged test loss results using Root Mean Square Error (RMSE) for each epoch for JAAD, while Figures \ref{fig:pie_test_loss} provide the corresponding test loss results using (RMSE) for PIE.

On the $JAAD_{pose}$ dataset, as shown in Table \ref{table_res_jaad}, SGNetPose+ outperforms the baseline across all metrics. It achieves an 8.5\% reduction in Mean Squared Error (MSE) for 15 frames (0.5 seconds ahead), a 6.8\% reduction for 30 frames (1 second ahead), and a 9.7\% reduction for 45 frames (1.5 seconds ahead). Furthermore, SGNetPose+ outperforms the baseline in the final frame prediction (FMSE) by 13.6\%, in centroid-based error (CMSE) by 12.4\%, and in the last frame's centroid prediction (CFMSE) by an impressive 16.9\%. These results highlight the model's ability to predict both short- and long-term pedestrian trajectories more accurately.

On the $PIE_{pose}$ dataset, SGNetPose+ with pose/skeleton data achieves notable improvements over the baseline model, reducing Mean Squared Error (MSE) by 7.3\% for 15 frames, 4.5\% for 30 frames, and 1.3\% for 45 frames. Additionally, when incorporating body angle data, SGNetPose+ outperforms the baseline on the final frame prediction (FMSE) by 2.6\%, on centroid-based error (CMSE) by 1.6\%, and on the last frame's centroid prediction (CFMSE) by 3.6\%. These results underscore the model's effectiveness in leveraging pose and body angle data for trajectory prediction.

\section{Analysis}

Predictions 1.5 seconds into the future exhibit greater uncertainty compared to those at 0.5 or 1 second, reflecting the inherent difficulty of forecasting further ahead, where even minor parameter adjustments can cause significant performance fluctuations. For example, in Figure \ref{fig:jaad_other_metrics}, the baseline model's FMSE metric jumps sharply from roughly 1700 to 2200, while in Figure \ref{fig:jaad_mse}, the MSE 15 metric shows smaller, more stable changes of about 10–20 points per epoch. A similar pattern is observed in the $PIE_{pose}$ dataset, where Figure \ref{fig:pie_other_metrics} shows a jump in the FMSE metric from 450 to 600 after 5 epochs, whereas the MSE 15 metric in Figure \ref{fig:pie_mse} remains relatively stable. These results highlight the compounding uncertainty in long-term predictions compared to shorter-term metrics.

Secondly, skeleton data enhance performance on the $JAAD_{pose}$ dataset, improving all metrics by 8\% to 16\%. On the PIE dataset, while skeleton data also outperform the baseline on metrics such as MSE 15, MSE 30, MSE 45, and CMSE, the improvements are more modest. This is likely due to the larger size of the PIE dataset, as shown in Table \ref{table_data_counts}, which may reduce the incremental benefit of the additional information.

Thirdly, body angle data underperform on the JAAD dataset but exceed the baseline across all metrics for the PIE dataset, though the gains are small. This result is noteworthy despite being derived indirectly via calculations, can still provide valuable insights. However, its effectiveness seems to be more pronounced with larger datasets, suggesting it requires sufficient data to fully realize its potential.

\section{Conclusion}

In this study, we introduced SGNetPose+, a robust enhancement of the SGNet model, which integrates pose information with bounding box data to improve pedestrian trajectory prediction in autonomous driving. The proposed architecture effectively captures complex spatial and temporal patterns. Extensive experiments on the $JAAD_{pose}$ and $PIE_{pose}$ datasets demonstrate the model's superior performance across various prediction metrics compared to baseline methods. These findings underscore the potential of integrating pose information for more accurate and reliable trajectory predictions.

\section{Future Work}

Future work on this topic could pursue several promising directions. Firstly, while SGNet was a foundational choice, it does not represent the current state-of-the-art in trajectory prediction. Recent models have demonstrated lower MSE results \cite{Lee2024GoalNetGA, Munir2024ContextawareML}. Testing the inclusion of pose data in these architectures, despite the challenges posed by the unavailability of their code, could offer valuable insights and further improve trajectory prediction performance.

Secondly, incorporating orientation as a pose feature offers potential for improvement. Beyond basic labels like left/right, finer classifications or angular variables, such as "towards the camera but slightly left/right," could capture subtle cues, possibly enriched with NLP techniques.

Thirdly, optimizing the selection of skeleton points could enhance the model's efficiency and accuracy. Instead of using all 13 points, as in this study, reducing the number to six or eight strategically selected points might enable the model to focus on the most informative features for motion prediction.

Lastly, 3D skeleton inference models could significantly improve accuracy by resolving overlapping points and providing precise body angles, making them particularly useful in dynamic, crowded scenarios where depth information is crucial.

{\small
\bibliographystyle{unsrt}
\bibliography{egbib}

\begin{thebibliography}{10}

\bibitem{alshami2024coool}
Ali~K AlShami, Ananya Kalita, Ryan Rabinowitz, Khang Lam, Rishabh Bezbarua, Terrance Boult, and Jugal Kalita.
\newblock Coool: Challenge of out-of-label a novel benchmark for autonomous driving.
\newblock {\em arXiv preprint arXiv:2412.05462}, 2024.

\bibitem{zhang2020research}
Hongjia Zhang, Yanjuan Liu, Chang Wang, Rui Fu, Qinyu Sun, and Zhen Li.
\newblock Research on a pedestrian crossing intention recognition model based on natural observation data.
\newblock {\em Sensors}, 20(6):1776, 2020.

\bibitem{WHO2023RoadSafety}
{World Health Organization (WHO)}.
\newblock {Global Status Report on Road Safety 2023}, 2023.
\newblock Accessed: 2024-12-15.

\bibitem{alshami2024smart}
Ali~K AlShami, Ryan Rabinowitz, Khang Lam, Yousra Shleibik, Melkamu Mersha, Terrance Boult, and Jugal Kalita.
\newblock Smart-vision: survey of modern action recognition techniques in vision.
\newblock {\em Multimedia Tools and Applications}, pages 1--72, 2024.

\bibitem{Li_2022_CVPR}
Lihuan Li, Maurice Pagnucco, and Yang Song.
\newblock Graph-based spatial transformer with memory replay for multi-future pedestrian trajectory prediction.
\newblock In {\em Proceedings of the IEEE/CVF Conference on Computer Vision and Pattern Recognition (CVPR)}, pages 2231--2241, June 2022.

\bibitem{Shi_2023_ICCV}
Liushuai Shi, Le~Wang, Sanping Zhou, and Gang Hua.
\newblock Trajectory unified transformer for pedestrian trajectory prediction.
\newblock In {\em Proceedings of the IEEE/CVF International Conference on Computer Vision (ICCV)}, pages 9675--9684, October 2023.

\bibitem{LIN2022111}
Tianyang Lin, Yuxin Wang, Xiangyang Liu, and Xipeng Qiu.
\newblock A survey of transformers.
\newblock {\em AI Open}, 3:111--132, 2022.

\bibitem{bock2017self}
Julian Bock, Till Beemelmanns, Markus Kl{\"o}sges, and Jens Kotte.
\newblock Self-learning trajectory prediction with recurrent neural networks at intelligent intersections.
\newblock In {\em VEHITS}, pages 346--351, 2017.

\bibitem{saleh2017intent}
Khaled Saleh, Mohammed Hossny, and Saeid Nahavandi.
\newblock Intent prediction of vulnerable road users from motion trajectories using stacked lstm network.
\newblock In {\em 2017 IEEE 20th International Conference on intelligent transportation systems (ITSC)}, pages 327--332. IEEE, 2017.

\bibitem{saleh2018intent}
Khaled Saleh, Mohammed Hossny, and Saeid Nahavandi.
\newblock Intent prediction of pedestrians via motion trajectories using stacked recurrent neural networks.
\newblock {\em IEEE Transactions on Intelligent Vehicles}, 3(4):414--424, 2018.

\bibitem{liu2020spatiotemporal}
Bingbin Liu, Ehsan Adeli, Zhangjie Cao, Kuan-Hui Lee, Abhijeet Shenoi, Adrien Gaidon, and Juan~Carlos Niebles.
\newblock Spatiotemporal relationship reasoning for pedestrian intent prediction.
\newblock {\em IEEE Robotics and Automation Letters}, 5(2):3485--3492, 2020.

\bibitem{rasouli2017they}
Amir Rasouli, Iuliia Kotseruba, and John~K Tsotsos.
\newblock Are they going to cross? a benchmark dataset and baseline for pedestrian crosswalk behavior.
\newblock In {\em Proceedings of the IEEE International Conference on Computer Vision Workshops}, pages 206--213, 2017.

\bibitem{liang2019peeking}
Junwei Liang, Lu~Jiang, Juan~Carlos Niebles, Alexander~G Hauptmann, and Li~Fei-Fei.
\newblock Peeking into the future: Predicting future person activities and locations in videos.
\newblock In {\em Proceedings of the IEEE/CVF conference on computer vision and pattern recognition}, pages 5725--5734, 2019.

\bibitem{yin2021multimodal}
Ziyi Yin, Ruijin Liu, Zhiliang Xiong, and Zejian Yuan.
\newblock Multimodal transformer networks for pedestrian trajectory prediction.
\newblock In {\em IJCAI}, pages 1259--1265, 2021.

\bibitem{alshami2023pose2trajectory}
Ali AlShami, Terrance Boult, and Jugal Kalita.
\newblock Pose2trajectory: Using transformers on body pose to predict tennis player’s trajectory.
\newblock {\em Journal of Visual Communication and Image Representation}, 97:103954, 2023.

\bibitem{piccoli2020fussi}
Francesco Piccoli, Rajarathnam Balakrishnan, Maria~Jesus Perez, Moraldeepsingh Sachdeo, Carlos Nunez, Matthew Tang, Kajsa Andreasson, Kalle Bjurek, Ria~Dass Raj, Ebba Davidsson, et~al.
\newblock Fussi-net: Fusion of spatio-temporal skeletons for intention prediction network.
\newblock In {\em 2020 54th asilomar conference on signals, systems, and computers}, pages 68--72. IEEE, 2020.

\bibitem{lorenzo2020rnn}
Javier Lorenzo, Ignacio Parra, Florian Wirth, Christoph Stiller, David~Fern{\'a}ndez Llorca, and Miguel~Angel Sotelo.
\newblock Rnn-based pedestrian crossing prediction using activity and pose-related features.
\newblock In {\em 2020 IEEE Intelligent Vehicles Symposium (IV)}, pages 1801--1806. IEEE, 2020.

\bibitem{kotseruba2021benchmark}
Iuliia Kotseruba, Amir Rasouli, and John~K Tsotsos.
\newblock Benchmark for evaluating pedestrian action prediction.
\newblock In {\em Proceedings of the IEEE/CVF winter conference on applications of computer vision}, pages 1258--1268, 2021.

\bibitem{fernando2018soft}
Tharindu Fernando, Simon Denman, Sridha Sridharan, and Clinton Fookes.
\newblock Soft+ hardwired attention: An lstm framework for human trajectory prediction and abnormal event detection.
\newblock {\em Neural networks}, 108:466--478, 2018.

\bibitem{huang2020long}
Zhe Huang, Aamir Hasan, Kazuki Shin, Ruohua Li, and Katherine Driggs-Campbell.
\newblock Long-term pedestrian trajectory prediction using mutable intention filter and warp lstm.
\newblock {\em IEEE Robotics and Automation Letters}, 6(2):542--549, 2020.

\bibitem{goldhammer2019intentions}
Michael Goldhammer, Sebastian K{\"o}hler, Stefan Zernetsch, Konrad Doll, Bernhard Sick, and Klaus Dietmayer.
\newblock Intentions of vulnerable road users—detection and forecasting by means of machine learning.
\newblock {\em IEEE transactions on intelligent transportation systems}, 21(7):3035--3045, 2019.

\bibitem{yao2021bitrap}
Yu~Yao, Ella Atkins, Matthew Johnson-Roberson, Ram Vasudevan, and Xiaoxiao Du.
\newblock Bitrap: Bi-directional pedestrian trajectory prediction with multi-modal goal estimation.
\newblock {\em IEEE Robotics and Automation Letters}, 6(2):1463--1470, 2021.

\bibitem{mangalam2020not}
Karttikeya Mangalam, Harshayu Girase, Shreyas Agarwal, Kuan-Hui Lee, Ehsan Adeli, Jitendra Malik, and Adrien Gaidon.
\newblock It is not the journey but the destination: Endpoint conditioned trajectory prediction.
\newblock In {\em Computer Vision--ECCV 2020: 16th European Conference, Glasgow, UK, August 23--28, 2020, Proceedings, Part II 16}, pages 759--776. Springer, 2020.

\bibitem{rasouli2019pie}
Amir Rasouli, Iuliia Kotseruba, Toni Kunic, and John~K Tsotsos.
\newblock Pie: A large-scale dataset and models for pedestrian intention estimation and trajectory prediction.
\newblock In {\em Proceedings of the IEEE/CVF International Conference on Computer Vision}, pages 6262--6271, 2019.

\bibitem{xu2022vitpose}
Yufei Xu, Jing Zhang, Qiming Zhang, and Dacheng Tao.
\newblock Vitpose: Simple vision transformer baselines for human pose estimation.
\newblock {\em Advances in Neural Information Processing Systems}, 35:38571--38584, 2022.

\bibitem{sidharta2024small}
Hanugra~Aulia Sidharta, Berlian Al~Kindhi, Eko~Mulyanto Yuniarno, and Mauridhi~Hery Purnomo.
\newblock Small group pedestrian crossing behaviour prediction using temporal angular 2d skeletal pose.
\newblock {\em Array}, page 100341, 2024.

\bibitem{wang2022stepwise}
Chuhua Wang, Yuchen Wang, Mingze Xu, and David~J Crandall.
\newblock Stepwise goal-driven networks for trajectory prediction.
\newblock {\em IEEE Robotics and Automation Letters}, 7(2):2716--2723, 2022.

\bibitem{Lee2024GoalNetGA}
Ching-Lin Lee, Zhi-Xuan Wang, Kuan-Ting Lai, and Amar Fadillah.
\newblock Goalnet: Goal areas oriented pedestrian trajectory prediction.
\newblock {\em ArXiv}, abs/2402.19002, 2024.

\bibitem{Munir2024ContextawareML}
Farzeen Munir and Tomasz~Piotr Kucner.
\newblock Context-aware multi-task learning for pedestrian intent and trajectory prediction.
\newblock {\em ArXiv}, abs/2407.17162, 2024.

\end{thebibliography}
}

\end{document}